\documentclass[]{spie}

\usepackage{amsmath,amsfonts,amssymb}
\usepackage{graphicx}
\usepackage{booktabs}
\usepackage[colorlinks=true, allcolors=blue]{hyperref}

\title{Topology-Informed Neural Networks for Flood Detection in Optical and Synthetic Aperture Radar Imagery}

\author[a]{Sophia Li}
\author[b]{Max Zhao*}
\author[a]{Raghu G. Raj}
\author[a]{Tianyu Chen}
\affil[a]{U.S. Naval Research Laboratory, Washington, DC, USA}
\affil[b]{Stanford University, Stanford, CA, USA}

\authorinfo{*E-mail:max.zhao@stanford.edu}

\begin{document}
\maketitle

\begin{abstract}
Floods frequently impact regions around the world. Rapid and accurate flood detection is crucial for emergency response and timely mitigation of human and economic loss. The expanding availability of satellite data and advances in artificial intelligence have enhanced monitoring of environmental hazards, but many flood events remain challenging to detect because cloud cover obscures optical satellite imagery. Rambour et al. introduced the SEN12-FLOOD dataset and extracted per-image features using a ResNet-50 convolutional neural network backbone, then fed these features into a gated recurrent unit network to show that temporal information can substantially improve accuracy compared to single-image baselines. More recently, Chamatidis et al. showed that a vision transformer can achieve strong performance with popular convolutional architectures. However, these models typically function as opaque black boxes, making it difficult to interpret their decision boundaries, learned features, and internal reasoning, especially in safety-critical domains like remote sensing. In contrast, topological data analysis (TDA) provides a mathematically grounded framework for capturing global structural features of data. TDA has emerged as a powerful tool for analyzing complex imagery, especially imagery with geometrically interpretable structures, of which floods are a prime candidate. In this work, we systematically evaluate topological descriptors for flood detection using the open-source SEN12-FLOOD dataset. By extracting topological features from each image and incorporating them into neural networks, we demonstrate that topological descriptors carry meaningful flood signals independently and complement existing networks to yield more robust and interpretable flood detection systems.
\end{abstract}


\section{Introduction}
Recently, the expanding availability of satellite data and rapid advancements in artificial intelligence have significantly enhanced our capability for monitoring environmental hazards \cite{Zhu_2017}. Nevertheless, the detection of many flood events remains challenging due to persistent cloud cover obscuring optical satellite imagery \cite{rs16040656}. Synthetic Aperture Radar (SAR) systems, such as the Sentinel-1 satellite, can penetrate clouds, but inherently contain speckle \cite{article}. As such, models that leverage both optical and SAR data can overcome individual sensor limitations.  
\\\\
A major issue in flood detection with SAR data is that many studies are tested on author-made datasets, which lack baselines and limit reproducibility \cite{rs16040656}. To address this issue, Rambour et al. \cite{isprs-archives-XLIII-B2-2020-1343-2020} developed the publicly available SEN12-FLOOD dataset, offering coregistered time series data from both Sentinel-1 (S1) SAR and Sentinel-2 (S2) multispectral imagery covering locations near 2018 and 2019 floods in Africa, Iran, and Australia. Both rural and urban areas are captured, as rural communities often lack resilient infrastructure and relief resources, while urban centers have concentrated populations and high-value assets. Each image in the sequence has a binary label specifying if a flood event has happened in the location of the sequence. This dataset leverages the two mentioned modalities of remote sensing data, and serves to assess the important role of raising flood alarms prompting human review and usage of more expensive flood segmentation algorithms, a crucial first step in the disaster relief process. As a baseline, Rambour et al. extracted per-image features using a ResNet-50 Convolutional Neural Network (CNN) \cite{DBLP:journals/corr/HeZRS15} backbone, then fed features into a Gated Recurrent Unit (GRU) \cite{cho2014learningphraserepresentationsusing} for flood classification. They demonstrated that temporal information substantially improves accuracy compared to single-image binary classifier baselines, with improvements up to 10\%. 
\\\\
Various experiments comparing different computer vision architectures have been conducted on this dataset. Mittal et al. \cite{inproceedings} surveyed the performance of various CNN architectures for binary classification, and Chamatidis et al. \cite{w16121670} showed that a Vision Transformers (ViT) can also achieve good performance. Nevertheless, Rambour's original findings highlight that time-series change detection is particularly well suited for flood detection tasks. For example, a river that appears narrow in one image but widens significantly in subsequent images may indicate flooding---information that cannot be reliably captured by static binary classifiers, especially given the existence of non-water frames in the dataset. Consequently, methods relying solely on binary classification of single images are unsuitable for practical deployment.
\\\\
Recently, Topological Data Analysis (TDA) has shown great promise as a tool in a variety of fields. The fact that TDA uses algebraic topology, means it can capture properties of complex data that traditional machine learning methods cannot capture \cite{coskunuzer2024topologicalmethodsmachinelearning}. Features capture global structural characteristics of a dataset that are unchanged under deformation, are provably stable under perturbation. As such, TDA shows promise for noisy and incomplete datasets. Another advantage of such a mathematically grounded framework is that it is contains human-understandable features, as opposed to most deep neural networks, whose features usually function as opaque black boxes. Such understanding is beneficial for safety-critical fields (e.g. disasters) where lives are at risk. Previous studies have demonstrated the success of combining convolutional and topological features for image classification across various domains, including medical images \cite{peng2023phgnetpersistenthomologyguided}, handwriting \cite{lima2023imageclassificationusingcombination}, and land data captured by optical remote sensing satellites \cite{sharma2025improvingremotesensingclassification}. Such success implies that convolutional (local textures) and topological (global structures) of complex datasets complement each other.
\\\\
In this work, we first improve on temporal-classification GRU baselines for SEN12-FLOOD by introducing transfer learning. Then, we propose a lightweight variant of a Gaussian topological embedding that achieves stable training convergence and improved performance when topological features are fed into a GRU, compared to the results in [4]. Finally, we demonstrate improvement when convolutional and topological features are both used, achieving a 98.9\% accuracy compared to a baseline of 95.7\%.

\section{Background}
\subsection{Dataset}
The SEN12-FLOOD dataset comprises a total of 335 sequences, each corresponding to a unique geographic location in Africa, Iran, or Australia. Each sequence contains multiple frames from S1 SAR and S2 multispectral optical sensors. On average, there are days or weeks between the capture of each frame and the full sequences can span months. The S1 frames were provided as two polarization channels Vertical transmit-Vertical receive (VV) and Vertical transmit-Horizontal receive (VH) in linear backscatter intensity, while the S2 frames included twelve spectral bands (B01, B02, B03, B04, B05, B06, B07, B08, B8A, B09, B11, B12 with pixel sizes 60m, 10m, 10m, 10m, 20m, 20m, 20m, 10m, 20m, 60m, 20m, 20m respectively). The S1 SAR dataset includes image frames acquired by both Sentinel-1A (S1A) and Sentinel-1B (S1B) satellites. As the two satellites follow distinct orbital paths, they observe the same regions from slightly different incidence angles, which may result in geometric variations within the imagery \cite{article}. The S2 data was inherited from the MediaEval 2019 dataset \cite{Bischke2019TheMS}. Each acquisition is annotated with a binary flood status label where $0$ denotes non-flooded and $1$ denotes flooded conditions. The dataset follows the hypothesis that flooded features persist in all post-flood observations of an area.  

\subsection{Topological Data Analysis}

We first describe the necessary topological constructions for a two-dimensional grayscale image and then introduce persistent homology, the main tool of TDA, which tracks the birth and death of features such as connected components and holes. These operations were implemented using the CubicalRipser Python libary \cite{DBLP:journals/corr/abs-2005-12692}.

\subsubsection*{Grid, hypercubes, and cubical complexes.} 

\textit{Intuition.} While TDA, especially on point clouds, conventionally uses simplices (triangles, tetrahedra, and higher dimensional analogs), it is more convenient for us to consider squares and cubes as our building blocks. Consider a $D$-dimensional integer grid. A cube is built by choosing, in each coordinate, either a unit interval (extending) or a single grid point (fixed). Faces result from turning an extended coordinate into a fixed one.
\\
\textit{Definition.} Fix $D\in\mathbb{N}$ and grid sizes $N_1,\dots,N_D\in\mathbb{N}$. A (unit) hypercube in $\mathbb{R}^D$ is specified by a pair $(v,\sigma)$ with $v\in\mathbb{Z}^D$ and $\sigma\in\{e,f\}^D$, represented by the product
\begin{equation}
Q(v,\sigma)\;=\;\prod_{i=1}^D
\begin{cases}
[v_i,\,v_i+1], & \sigma_i = e, \\[2pt]
\{v_i\}, & \sigma_i = f.
\end{cases}
\label{eq:q_v_sigma}
\end{equation}
Its dimension is $\dim Q=\#\{i:\sigma_i=e\}$. For any index $k$ with $\sigma_k=e$, $Q$ has two codimension-1 faces in the $k$th direction:
\begin{equation}
\text{lower face: } Q\bigl(v,\,\sigma\text{ with }\sigma_k\!\leftarrow\! f\bigr),\qquad
\text{upper face: } Q\bigl(v+e_k,\,\sigma\text{ with }\sigma_k\!\leftarrow\! f\bigr)
\label{eq:faces}
\end{equation}
A cubical complex, $\mathcal{C}$, is a finite set of such hypercubes closed under faces, meaning that whenever a cube, $Q$, belongs to $\mathcal{C}$, all of its faces (as in \eqref{eq:faces}) must also belong to $\mathcal{C}$.

\subsubsection*{2D grayscale maps as cubical complexes (V-construction).}
\textit{Intuition.} Each pixel is treated as a vertex, with edges and squares included as needed. Collectively, these vertices, edges, and squares form the cubical complex.
\\
\textit{Definition.} For $D=2$, let the vertex index set be $\{0,\dots,N_1-1\}\times\{0,\dots,N_2-1\}$. A grayscale image is a function on vertices
\begin{equation}
v:\ \{\text{0-cubes}\}\to\mathbb{R}_{\ge 0} ,\qquad v(i,j)=\text{intensity at vertex }(i,j)
\label{eq:v_intensity}
\end{equation}
The cubical complex consists of all unit edges and unit squares whose vertices lie in this index set; that is, each unit $k$-cube is included precisely when all of its vertices are contained in the index set.  \\
\\
\textit{Remark.} The alternative T-construction assigns intensities to top cells and extends them to faces via the $\min$ operation. The V- and T-constructions are related by duality. In our case, the particular choice of construction does not substantially affect the results.

\subsubsection*{Lower-star extension and filtration (vertex-based).}
\textit{Intuition.} Turn on all cells whose vertices are at most threshold $t$; as $t$ increases, the complex grows.
\\
\textit{Definition.} Extend $v$ from vertices to all cubes by
\begin{equation}
f(c)\;=\;\max\bigl\{\,v(p)\;:\;p\text{ is a vertex of }c\,\bigr\}
\label{eq:f_c}
\end{equation}
Then $f$ is monotone under inclusion: if $c\subseteq c'$ then $f(c)\le f(c')$. For $t\in\mathbb{R}$, define the sublevel-set filtration
\begin{equation}
C_t\;=\;\bigl\{\,c\in\mathcal{C}\ :\ f(c)\le t\,\bigr\}
\label{eq:C_t}
\end{equation}
This yields a nested sequence $t\le s\Rightarrow C_t\subseteq C_s$ that grows from a low-intensity structure to the full complex.

\subsubsection*{Persistent homology.}
Let $\Bbbk$ be a field (e.g. $\mathbb{Z}_2$). Given the sublevel-set filtration $\{C_t\}_{t \in \mathbb{R}}$ from the previous section, the homology groups $H_k(C_t;\Bbbk)$ summarize $k$-dimensional topological features present at threshold $t$: connected components ($k{=}0$), loops ($k{=}1$), and higher-dimensional voids for $k \ge 2$. In our two-dimensional case, we only track $k{=}0$ and $k{=}1$. As $t$ increases, a feature is born when it first appears in $H_k(C_t)$ and dies when it merges into an older feature (for $k{=}0$) or becomes filled (for $k{=}1$). Persistent homology records these events as birth--death pairs $(b_i,d_i)$.  
\\
\textit{Remark.} For $k=0$, it matters which component is considered older, because by convention, the younger component (with the larger birth time) is always the one that dies. This convention ensures that each connected component has a unique birth and death time, making the pairing well-defined.

\paragraph{Persistence diagrams (PDs).}
For each $k\ge 0$, the $k$-dimensional persistence diagram is the multiset
\begin{equation}
\mathrm{Dgm}_k\;=\;\bigl\{(b_i,d_i)\in\mathbb{R}^2\cup\{\infty\}\bigr\}
\label{eq:dgm_k}
\end{equation}
with one point per $k$-feature. All points must exist above the diagonal $b{=}d$, as features cannot die before they are born. Points far from the diagonal correspond to persistent features with long lifetimes, $d_i-b_i$, reflecting stable, meaningful topological structures. Points near the diagonal have short lifetimes and are typically sensitive to small perturbations, representing noise rather than significant structures. Points with death threshold infinity, or features that persisted through the filtration, are nonnegligible. Persistence diagrams are stable: small perturbations in the input function result in only small changes in the diagrams under the bottleneck distance~\cite{cohensteiner2007stability}.

\section{Methods}
\subsection{Preprocessing}

For \textbf{Sentinel-1}, VV and VH polarization channels were initially supplied in linear units after radiometric calibration and range Doppler terrain correction. Valid (positive, finite) pixels were converted into deciBel (dB) scale using a logarithmic transformation,
\begin{equation}
\sigma^0_{\mathrm{dB}} = 10 \cdot \log_{10} \left( \max(\sigma^0, \epsilon) \right)
\label{eq:sigma_dB}
\end{equation}
with $\epsilon = 10^{-6}$ to avoid numerical underflow. Then, a log-scaled grayscale map was generated per acquisition by aggregating both channels as
\begin{equation}
I_{\mathrm{gray}} = 10 \cdot \log_{10}(VV + VH)
\label{eq:igray}
\end{equation}
Acquisitions with missing bands, insufficient valid pixels ($<75\%$), or negligible intensity standard deviation ($<0.001$) were discarded.
\\\\
For \textbf{Sentinel-2}, images had level 2A atmospheric correction. We first radiometrically scaled stacks by dividing all values by $10{,}000$ yielding normalized surface reflectance values in the range $[0,1]$. To ensure geometric consistency, each band was resampled by bilinear interpolation to match the spatial resolution of the reference band (B03 or Green, 10m size). Then, a grayscale map on the Normalized Difference Water Index (NDWI) \cite{GAO1996257} was generated using the Green (B03) and Near-Infrared (B08) bands 
\begin{equation}
NDWI = \frac{B03 - B08}{B03 + B08 + \epsilon}
\label{eq:ndwi}
\end{equation}
clipped $[-1,1]$ and negated for watered regions to have lower intensity values (as consistent with the S1 grayscale maps). Acquisitions with missing bands, insufficient valid pixels ($<75\%$), or negligible intensity standard deviation ($<0.001$) were discarded.
\\\\
Stacked bands were saved and resized to 120x120 for transfer CNN learning, and grayscale maps were used for topological operations. Following the pretrained models' convention, we dropped the 60m resolution bands, B01 and B09 for S2 acquisitions, saving a 10-channel stack. After preprocessing, there were 3418 S1 images, 1325 of which were flooded, and 1983 S2 images, 446 of which were flooded. 192 out of the 335 sequences contained a flood event. We used the given split of 267 train and 68 test sequences. It is important to note that, due to preprocessing, when training S1-only models, some sequences had zero valid S1 acquisitions yielding 250 train and 62 test sequences. On average, there were 10.20 S1 acquisitions per sequence and 5.92 S2 acquisitions per sequence. Preprocessing was done using the RasterIO \cite{gillies_2019} Python library.

\subsection{Gaussian Embedding of Persistence Diagrams}

As part of the intended raw nature of the dataset, optical images sometime contain cloud cover and have variations in atmospheric conditions across days, while SAR images contain speckle and geometric distortions due to incidence angle variety.  Thus, when using raw persistence diagrams, equivalent representations (Betti curves, persistence barcodes), or scalar summaries (amplitude, entropy, Wasserstein distances), our topological features were not able to converge when training and did not carry meaning across different days. We introduce a lightweight Gaussian embedding of persistence diagrams that allows convergence and fast training.

\subsubsection*{Vectorizations of persistence diagrams}

Persistence diagrams are finite multisets in the birth--death plane and do not naturally inhabit a Hilbert space. Many learning algorithms, however, require fixed-length, differentiable, and stable features. Several vectorization schemes have been proposed to bridge this gap. Persistence landscapes map PDs to functional representations that can be discretized on a grid \cite{bubenik2015landscapes}. Persistence images smooth PDs with Gaussian kernels and integrate the resulting surface over pixels to obtain a finite-dimensional vector \cite{adams2017persistence}. Kernel methods operate directly on diagrams, including the persistence scale-space kernel \cite{reininghaus2015multiscale}, the persistence-weighted Gaussian kernel \cite{kusano2016pwgk}, and the sliced Wasserstein kernel \cite{carriere2017sliced}. All of these approaches build on the fundamental stability of persistence diagrams under perturbations \cite{cohensteiner2007stability}, and Gaussian or heat-kernel-based embeddings, in particular, inherit Lipschitz-type stability with respect to bottleneck and Wasserstein distances.

\paragraph{The embedding}
Given a PD $D=\{(b_i,d_i)\}$ with lifetimes $\ell_i=\max(d_i-b_i,0)$, for each homology dimension, $k$, and modality, $m$, we define a grid of $G\times G$ centers $\mathcal{C}^{(m)}_k=\{c_j\}_{j=1}^{G^2}$ by empirical birth and death quantiles computed on the training split. With a bandwidth $\sigma^{(m)}_k>0$ estimated from pooled birth and death scales, we define
\[
\phi^{(m,k)}_j(D)=\sum_{(b_i,d_i)\in D_k}\!\ell_i\,
\exp\!\Big(-\tfrac{1}{2(\sigma^{(m)}_k)^2}\,\Vert (b_i,d_i)-c_j\Vert_2^2\Big),
\quad
\Phi^{(m,k)}(D)=(\phi^{(m,k)}_j)_{j=1}^{G^2}.
\]
We concatenate $\Phi$ for $k=0$ and $k=1$ to form a vector of size $2G^2$. We trim and only keep the top points by lifetime per $k$ to reduce cost and only retain the most important features. 
\\\\
This construction is equivalent to persistence images \cite{adams2017persistence} but uses quantile-adaptive grids to help mitigate scale shifts. As with Gaussian heat-based summaries \cite{reininghaus2015multiscale}, the map is differentiable and inherits Lipschitz-type stability from PD stability \cite{cohensteiner2007stability}. We avoid kernels such as \cite{reininghaus2015multiscale,kusano2016pwgk,carriere2017sliced} because they are contingent on a pairwise similarity assumption across the whole dataset: as mentioned before, frames showing water aren't necessarily flooded, so an explicit vectorization is better for the change detection task.

\paragraph{Practical procedure}
For each homology dimension $k\in\{0,1\}$ and modality $m$, we collect births and deaths from the training split, $\mathcal{T}$, and form a $G\times G$ grid of centers $\mathcal{C}^{(m)}_k=\{c^{(m,k)}_j\}_{j=1}^{G^2}$ using empirical quantiles along birth and death. A bandwidth, $\sigma^{(m)}_k$, is set from pooled birth and death scales on $\mathcal{T}$. Given a diagram, $D$, with lifetimes $\ell_i=\max(d_i-b_i,0)$, we compute for each grid center
\[
\phi^{(m,k)}_j(D)=\sum_{(b_i,d_i)\in D_k}\!\ell_i\,\exp\!\Big(-\tfrac{\| (b_i,d_i)-c^{(m,k)}_j\|_2^2}{2(\sigma^{(m)}_k)^2}\Big),
\]
assemble $\Phi^{(m,k)}(D)=(\phi^{(m,k)}_j)_{j=1}^{G^2}$, and concatenate $k=0,1$ to obtain a $2G^2$-dimensional vector. In practice, we keep the top points by lifetime per $k$ to reduce cost. Centers and bandwidths are fitted once on $\mathcal{T}$ and reused for validation and test.

\subsection{Feature Extraction and Training}\label{sec:feature-train}

\begin{figure}[ht]
  \centering
  \includegraphics[width=\linewidth]{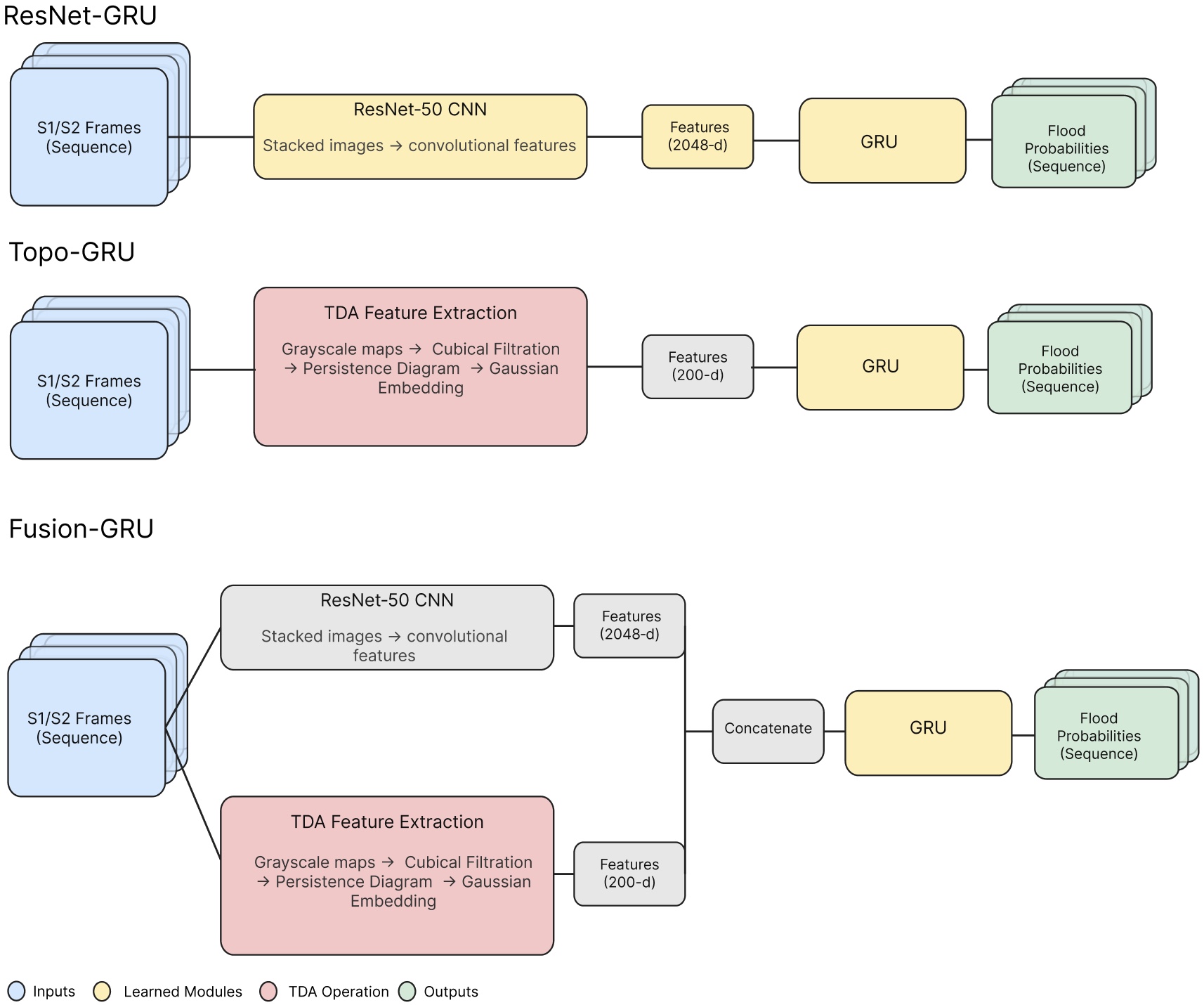}
  \caption{Diagram of ResNet50-GRU model, Topological (Topo)-GRU model and Joint (Fusion-GRU) model.}
  \label{fig:models-diagram}
\end{figure}

We propose and evaluate three sequence models that map variable-length sequences to per-frame flood probabilities: (i) Topological (Topo)-GRU, (ii) ResNet50-GRU, and (iii) Joint (Fusion-GRU) Model, as in Figure~\ref{fig:models-diagram}. All pipelines are implemented in Python with PyTorch \cite{paszke2019pytorchimperativestylehighperformance}, and are trained three times: on S1 data only, S2 data only, and the dual dataset that merges S1 and S2 as input.

\subsubsection*{Topological Features Only (Topo-GRU)}
\paragraph{Feature extraction (sublevel-set cubical filtration $\rightarrow$ PDs $\rightarrow$ Gaussian embedding, single modality).}
Given grayscale frames from modality S1 or S2, we construct PDs as in Section~2.2 and retain the top $K{=}200$ points per homology dimension by lifetime. Each diagram is mapped to a fixed-length vector via the Gaussian grid embedding defined as:
\begin{equation}
\Phi(D)\;=\;\operatorname{concat}\big(\Phi^{(0)}(D),\,\Phi^{(1)}(D)\big),
\label{eq:Phi_D}
\end{equation}
with $\phi^{(k)}_j$ computed from the centers and bandwidths fitted once on $\mathcal{T}$ and reused unchanged for validation and test. We use $G=10$ and $\lambda=0.1$ yielding output dimension $2G^2=200$. PDs are computed with CubicalRipser~\cite{DBLP:journals/corr/abs-2005-12692}, and parameters are maintained separately for S1 and S2.

\paragraph{Model.}
A GRU with hidden size $H{=}256$ consumes $\Phi(D)$ at each time step. Unlike previous work that used a binary classification head, our last layer (size one) produces a per-step probability. Specifically, the scalar logit, $z_t$, is an affine transformation of the hidden state, and a sigmoid activation, $\sigma(z_t)$, returns the frame-level flood probability.

\subsubsection*{Convolutional Features Only (ResNet50-GRU)}
\paragraph{Feature extraction.}
As per suggestion from the dataset authors \cite{isprs-archives-XLIII-B2-2020-1343-2020}, we initialize from ResNet-50 encoders pretrained on BigEarthNet \cite{hackel2024configilm, clasen2024refinedbigearthnet} S1 and S2 land classification datasets and discard any task-specific heads from pretraining. The first convolution was already modified to accept 2 or 10 channels. Given a $2\times120\times120$ or $10\times120\times120$ frame, the encoder applies global average pooling to the final feature maps producing a 2048-dimensional descriptor, $\psi_t$.
 
\paragraph{Model.}
The GRU head structure is the same as above for Topo-GRU. For the dual dataset, a 16-dimensional, learned embedding representing the modality is appended to $\psi_t$ before the GRU, resulting in vector size $2064$. The encoder is unfrozen during training.

\subsubsection*{Fusion of Features (Fusion-GRU)}
\paragraph{Feature extraction (late concatenation).}
For each frame, obtain the topological descriptor, $\phi(D)$, and the ResNet-50 descriptor, $\psi_t$, by passing through the trained GRU encoder for the frame's modality yielding a 2248-dimensional descriptor. The encoder is frozen during this process, and only the GRU is trained.

\paragraph{Model.}
The GRU head structure is the same as for Topo-GRU. For the dual dataset, a 16-dimensional, learned embedding representing the modality is appended to $\Phi$ before the GRU resulting in vector size $2264$. 

\subsubsection*{Common training protocol}
\paragraph{Training.}
We use class-weighted logistic loss
\begin{equation}
\mathcal{L}=-\frac{1}{N}\sum_{i=1}^{N}\Big[w_+\,y_i\log\sigma(z_i)+(1-y_i)\log\big(1-\sigma(z_i)\big)\Big],\qquad
w_+=\frac{1-\pi}{\pi}
\label{eq:loss_weighted}
\end{equation}
where $\pi$ is the positive rate over valid training frames, $y_i$ is the true label, and $\sigma(z_i)$ is the output probability. The classifier bias is initialized to $\log\frac{\pi}{1-\pi}$. All models are trained with the Adam optimizer, learning rate $0.001$, batch size $8$, and a maximum of $200$ epochs. To handle varying sequence length when batching, sequences are padded to the longest length in the batch; masks exclude padding from both the loss and all evaluation metrics.

\paragraph{Evaluation Metrics.}
During training, model performance is assessed using standard classification metrics derived from the confusion matrix using probability threshold $\tau=0.5$: true positives (TP), true negatives (TN), false positives (FP), and false negatives (FN). Standard evaluation metrics are listed:

\begin{itemize}
  \item \textbf{Accuracy}: $\displaystyle Acc = \frac{TP + TN}{TP + TN + FP + FN}$
  \item \textbf{Precision}: $\displaystyle P = \frac{TP}{TP + FP}$
  \item \textbf{Recall}: $\displaystyle R = \frac{TP}{TP + FN}$
  \item \textbf{F$_1$-score}: $\displaystyle F_1 = \frac{2PR}{P + R}$
  \item \textbf{F$_\beta$-score}: $\displaystyle F_{\beta} = \frac{(1+\beta^2)PR}{\beta^2 P + R}$
\end{itemize}

\paragraph{Model selection and early stopping.}
Accuracy by itself is not an adequate metric for assessing deployability, particularly when using imbalanced datasets where non-flooded frames outnumber flooded ones. Given class imbalance, we do not select best training models on accuracy, as done in Rambour~\cite{isprs-archives-XLIII-B2-2020-1343-2020}. Instead, we monitor a recall-tilted $F_\beta$ with $\beta=\sqrt{2}$, which upweights recall twofold relative to precision. In other words, we consider missing a flood twice as bad as raising a false alarm. Early stopping halts training after 20 epochs without validation $F_\beta$ improvement.

\paragraph{Model deployment threshold.}
Post-training, we sweep the positive decision threshold $\tau\in[0,1]$ for probabilities and choose $\tau^\star$ for maximal recall subject to precision $\ge 0.90$. If no $\tau$ satisfies this constraint, we fall back to the threshold with maximum $F_1$.

\section{Results and Discussion}\label{sec:results}

\subsection{Results}

\begin{table}[ht!]
    \centering
    \renewcommand{\arraystretch}{1.2}
    \caption{Comparison of GRU model performance on S1, S2, and dual-modality tasks. We report $F_{\beta}$, $F_1$-score,  accuracy, previously reported baseline accuracy by \cite{isprs-archives-XLIII-B2-2020-1343-2020}, and best recall achieved at $90\%$ precision.}
    \begin{tabular}{lcccccc}
        \toprule
        \textbf{Model} & $\boldsymbol{F_{\beta}}$ & $\boldsymbol{F_{1}}$ & \textbf{Acc} & \textbf{Prev Acc} & \textbf{R@90\%P} \\
        \midrule
        ResNet50-GRU S1      & 0.944 & 0.941 & 0.958 & 0.875 & 0.954 \\
        Topo-GRU S1          & 0.903 & 0.886 & 0.914 &   --   & 0.713 \\
        ResNet50-GRU S2      & 0.976 & 0.971 & 0.988 & 0.930 & 0.988 \\
        Topo-GRU S2          & 0.892 & 0.897 & 0.958 &   --   & 0.881 \\
        ResNet50-GRU Dual    & 0.947 & 0.954 & 0.974 & 0.957 & 0.953 \\
        Topo-GRU Dual        & 0.916 & 0.904 & 0.941 &   --   & 0.903 \\
        \textbf{Fusion-GRU Dual}      & 0.980 & 0.982 & 0.989 &   --   & 0.986 \\
        \bottomrule
    \end{tabular}
    
    \label{tab:results}
\end{table}

\begin{table}[ht!]
    \centering
    \renewcommand{\arraystretch}{1.2}
    \caption{Comparison of GRU models on S1, S2, and dual-modality tasks with TP, FP, TN and FN.}
    \begin{tabular}{lcccc}
        \toprule
        \textbf{Model} & \textbf{TP} & \textbf{FP} & \textbf{TN} & \textbf{FN} \\
        \midrule
        ResNet50-GRU S1   & 185 & 13 & 336 & 10 \\
        Topo-GRU S1     & 183 & 35 & 314 & 12 \\
        ResNet50-GRU S2   & 83  & 4  & 313 & 1  \\
        Topo-GRU S2     & 74  & 7  & 310 & 10 \\
        ResNet50-GRU Dual & 260 & 6  & 660 & 19 \\
        Topo-GRU Dual   & 263 & 40 & 626 & 16 \\
        \textbf{Fusion-GRU Dual}    & 272 & 3 & 663 &7 \\
        \bottomrule
    \end{tabular}
    
    \label{tab:uni_confusion}
\end{table}

\subsection{Discussion} 

The ResNet-50 GRUs transfer-learned from BigEarthNet encoders achieve benchmark performance gains of 8.3\%, 5.8\%, and 1.7\% for the S1, S2, and Dual tasks, respectively. The largest improvement from exposure to the larger dataset for pretraining occurs with S1 SAR data, as unlike optical data, SAR is not conventionally used as input for CNNs. Compared to ResNet50-GRU models, which achieve 95.8\% accuracy on S1 and 98.8\% on S2, our Topo-GRUs perform slightly worse as expected, but still attain solid results of 91.4\% and 95.8\%, respectively, while using only about one-tenth of the feature size. Our Fusion-GRU Dual model achieves the best results, corroborating the statement that topological and convolutional features complement each other.

These results mark the first application of TDA directly to images for flood detection, and the first use of persistent homology on SAR imagery showing promise for further exploration. Future directions include investigating more advanced feature-combination strategies such as attention mechanisms, integrating persistence diagrams or persistence images into CNNs \cite{peng2023phgnetpersistenthomologyguided}, and allowing the topological feature embeddings to remain unfrozen and learnable. For example, Perslay \cite{carriere2020perslayneuralnetworklayer} enables optimization of centers and standard deviations, and alternative radial basis functions for embeddings could also be explored \cite{JMLR:v20:18-358}.

\acknowledgments
This work was funded by the Office of Naval Research under the Naval Research Laboratory (NRL) Base Program and the NRL Science and Engineering Apprenticeship Program (SEAP).

\newpage
\bibliography{references}

@article{Zhu_2017,
  author  = {Zhu, Xiao Xiang and Tuia, Devis and Mou, Lichao and Xia, Gui-Song and Zhang, Liangpei and Xu, Feng and Fraundorfer, Friedrich},
  title   = {Deep Learning in Remote Sensing: A Comprehensive Review and List of Resources},
  journal = {IEEE Geoscience and Remote Sensing Magazine},
  volume  = {5},
  number  = {4},
  pages   = {8--36},
  year    = {2017},
  doi     = {10.1109/MGRS.2017.2762307}
}

@article{rs16040656,
  author  = {Amitrano, Donato and Di Martino, Gerardo and Di Simone, Alessio and Imperatore, Pasquale},
  title   = {Flood Detection with {SAR}: A Review of Techniques and Datasets},
  journal = {Remote Sensing},
  volume  = {16},
  number  = {4},
  pages   = {656},
  year    = {2024},
  doi     = {10.3390/rs16040656}
}

@article{article,
  author  = {Moreira, Alberto and Prats-Iraola, Pau and Younis, Marwan and Krieger, Gerhard and Hajnsek, Irena and Papathanassiou, Konstantinos P.},
  title   = {A Tutorial on Synthetic Aperture Radar},
  journal = {IEEE Geoscience and Remote Sensing Magazine},
  volume  = {1},
  number  = {1},
  pages   = {6--43},
  year    = {2013},
  doi     = {10.1109/MGRS.2013.2248301}
}

@inproceedings{isprs-archives-XLIII-B2-2020-1343-2020,
  author    = {Rambour, C. and Audebert, N. and Koeniguer, E. and Le Saux, B. and Crucianu, M. and Datcu, M.},
  title     = {Flood Detection in Time Series of Optical and {SAR} Images},
  booktitle = {The International Archives of the Photogrammetry, Remote Sensing and Spatial Information Sciences},
  volume    = {XLIII-B2-2020},
  pages     = {1343--1346},
  year      = {2020},
  doi       = {10.5194/isprs-archives-XLIII-B2-2020-1343-2020}
}

@article{DBLP:journals/corr/HeZRS15,
  author        = {He, Kaiming and Zhang, Xiangyu and Ren, Shaoqing and Sun, Jian},
  title         = {Deep Residual Learning for Image Recognition},
  journal       = {CoRR},
  volume        = {abs/1512.03385},
  year          = {2015},
  eprint        = {1512.03385},
  archivePrefix = {arXiv},
  url           = {https://arxiv.org/abs/1512.03385}
}

@misc{cho2014learningphraserepresentationsusing,
  author        = {Cho, Kyunghyun and van Merrienboer, Bart and Gulcehre, Caglar and Bahdanau, Dzmitry and Bougares, Fethi and Schwenk, Holger and Bengio, Yoshua},
  title         = {Learning Phrase Representations using {RNN} Encoder--Decoder for Statistical Machine Translation},
  year          = {2014},
  eprint        = {1406.1078},
  archivePrefix = {arXiv},
  url           = {https://arxiv.org/abs/1406.1078}
}

@inproceedings{inproceedings,
  author    = {Mittal, Bhavuk and Vanzara, Jui and Sajidha, S. A.},
  title     = {Navigating Spatial Insights: Comparative Exploration of Deep Learning Algorithms for Flood Detection using Remote Sensing Data},
  booktitle = {2024 IEEE International Conference on Interdisciplinary Approaches in Technology and Management for Social Innovation ({IATMSI})},
  pages     = {1--6},
  year      = {2024},
  doi       = {10.1109/IATMSI60426.2024.10502981}
}

@article{w16121670,
  author  = {Chamatidis, Ilias and Istrati, Denis and Lagaros, Nikos D.},
  title   = {Vision Transformer for Flood Detection Using Satellite Images from {Sentinel-1} and {Sentinel-2}},
  journal = {Water},
  volume  = {16},
  number  = {12},
  pages   = {1670},
  year    = {2024},
  doi     = {10.3390/w16121670}
}

@misc{coskunuzer2024topologicalmethodsmachinelearning,
  author        = {Coskunuzer, Baris and Akcora, Cuneyt Gurcan},
  title         = {Topological Methods in Machine Learning: A Tutorial for Practitioners},
  year          = {2024},
  eprint        = {2409.02901},
  archivePrefix = {arXiv},
  url           = {https://arxiv.org/abs/2409.02901}
}

@inproceedings{peng2023phgnetpersistenthomologyguided,
  author    = {Peng, Yaopeng and Wang, Hongxiao and Sonka, Milan and Chen, Danny Z.},
  title     = {{PHG-Net}: Persistent Homology Guided Medical Image Classification},
  booktitle = {Proceedings of the IEEE/CVF Winter Conference on Applications of Computer Vision ({WACV})},
  pages     = {7583--7592},
  year      = {2024}
}

@misc{lima2023imageclassificationusingcombination,
  author        = {Lima, Mariana D{\'o}ria Prata and Giraldi, Gilson Antonio and Miranda Junior, Gast{\~a}o Flor{\^e}ncio},
  title         = {Image Classification using Combination of Topological Features and Neural Networks},
  year          = {2023},
  eprint        = {2311.06375},
  archivePrefix = {arXiv},
  url           = {https://arxiv.org/abs/2311.06375}
}

@misc{sharma2025improvingremotesensingclassification,
  author        = {Sharma, Aaryam},
  title         = {Improving Remote Sensing Classification using Topological Data Analysis and Convolutional Neural Networks},
  year          = {2025},
  eprint        = {2507.10381},
  archivePrefix = {arXiv},
  url           = {https://arxiv.org/abs/2507.10381}
}

@inproceedings{Bischke2019TheMS,
  author    = {Bischke, Benjamin and Helber, Patrick and Brugman, Simon and Basar, Erkan and Zhao, Zhengyu and Larson, Martha and Pogorelov, Konstantin},
  title     = {The Multimedia Satellite Task at {MediaEval} 2019},
  booktitle = {MediaEval Benchmarking Initiative for Multimedia Evaluation},
  year      = {2019}
}

@article{DBLP:journals/corr/abs-2005-12692,
  author        = {Kaji, Shizuo and Sudo, Takeki and Ahara, Kazushi},
  title         = {Cubical Ripser: Software for Computing Persistent Homology of Image and Volume Data},
  journal       = {CoRR},
  volume        = {abs/2005.12692},
  year          = {2020},
  eprint        = {2005.12692},
  archivePrefix = {arXiv},
  url           = {https://arxiv.org/abs/2005.12692}
}

@article{cohensteiner2007stability,
  author  = {Cohen-Steiner, David and Edelsbrunner, Herbert and Harer, John},
  title   = {Stability of Persistence Diagrams},
  journal = {Discrete \& Computational Geometry},
  volume  = {37},
  number  = {1},
  pages   = {103--120},
  year    = {2007},
  doi     = {10.1007/s00454-006-1276-5}
}

@article{GAO1996257,
  author  = {Gao, Bo-Cai},
  title   = {{NDWI}---A Normalized Difference Water Index for Remote Sensing of Vegetation Liquid Water from Space},
  journal = {Remote Sensing of Environment},
  volume  = {58},
  number  = {3},
  pages   = {257--266},
  year    = {1996},
  doi     = {10.1016/S0034-4257(96)00067-3}
}

@misc{gillies_2019,
  author = {Gillies, Sean and others},
  title  = {Rasterio: Geospatial Raster {I/O} for {Python} Programmers},
  year   = {2013--},
  url    = {https://rasterio.readthedocs.io/}
}

@article{bubenik2015landscapes,
  author  = {Bubenik, Peter},
  title   = {Statistical Topological Data Analysis using Persistence Landscapes},
  journal = {Journal of Machine Learning Research},
  volume  = {16},
  number  = {3},
  pages   = {77--102},
  year    = {2015}
}

@article{adams2017persistence,
  author  = {Adams, Henry and Emerson, Tegan and Kirby, Michael and Neville, Rachel and Peterson, Chris and Shipman, Patrick and Chepushtanova, Sofya and Hanson, Eric and Motta, Francis and Ziegelmeier, Lori},
  title   = {Persistence Images: A Stable Vector Representation of Persistent Homology},
  journal = {Journal of Machine Learning Research},
  volume  = {18},
  number  = {8},
  pages   = {1--35},
  year    = {2017}
}

@inproceedings{reininghaus2015multiscale,
  author    = {Reininghaus, Jan and Huber, Stefan and Bauer, Ulrich and Kwitt, Roland},
  title     = {A Stable Multi-Scale Kernel for Topological Machine Learning},
  booktitle = {Proceedings of the IEEE Conference on Computer Vision and Pattern Recognition ({CVPR})},
  pages     = {4741--4748},
  year      = {2015}
}

@inproceedings{kusano2016pwgk,
  author    = {Kusano, Genki and Fukumizu, Kenji and Hiraoka, Yasuaki},
  title     = {Persistence Weighted Gaussian Kernel for Topological Data Analysis},
  booktitle = {Proceedings of the 33rd International Conference on Machine Learning},
  series    = {Proceedings of Machine Learning Research},
  volume    = {48},
  pages     = {2004--2013},
  year      = {2016},
  publisher = {PMLR}
}

@inproceedings{carriere2017sliced,
  author    = {Carri{\`e}re, Mathieu and Cuturi, Marco and Oudot, Steve},
  title     = {Sliced Wasserstein Kernel for Persistence Diagrams},
  booktitle = {Proceedings of the 34th International Conference on Machine Learning},
  series    = {Proceedings of Machine Learning Research},
  pages     = {664--673},
  year      = {2017},
  publisher = {PMLR}
}

@inproceedings{paszke2019pytorchimperativestylehighperformance,
  author    = {Paszke, Adam and Gross, Sam and Massa, Francisco and Lerer, Adam and Bradbury, James and Chanan, Gregory and Killeen, Trevor and Lin, Zeming and Gimelshein, Natalia and Antiga, Luca and Desmaison, Alban and K{"o}pf, Andreas and Yang, Edward and DeVito, Zachary and Raison, Martin and Tejani, Alykhan and Chilamkurthy, Sasank and Steiner, Benoit and Fang, Lu and Bai, Junjie and Chintala, Soumith},
  title     = {{PyTorch}: An Imperative Style, High-Performance Deep Learning Library},
  booktitle = {Advances in Neural Information Processing Systems 32},
  pages     = {8024--8035},
  year      = {2019}
}

@article{hackel2024configilm,
  author  = {Hackel, Leonard and Clasen, Kai Norman and Demir, Beg{\"u}m},
  title   = {{ConfigILM}: A General Purpose Configurable Library for Combining Image and Language Models for Visual Question Answering},
  journal = {SoftwareX},
  volume  = {26},
  pages   = {101731},
  year    = {2024},
  doi     = {10.1016/j.softx.2024.101731}
}

@misc{clasen2024refinedbigearthnet,
  author        = {Clasen, Kai Norman and Hackel, Leonard and Burgert, Tom and Sumbul, Gencer and Demir, Beg{\"u}m and Markl, Volker},
  title         = {{reBEN}: Refined {BigEarthNet} Dataset for Remote Sensing Image Analysis},
  year          = {2024},
  eprint        = {2407.03653},
  archivePrefix = {arXiv},
  url           = {https://arxiv.org/abs/2407.03653}
}

@inproceedings{carriere2020perslayneuralnetworklayer,
  author    = {Carri{\`e}re, Mathieu and Chazal, Fr{\'e}d{\'e}ric and Ike, Yuichi and Lacombe, Th{\'e}o and Royer, Martin and Umeda, Yuhei},
  title     = {{PersLay}: A Neural Network Layer for Persistence Diagrams and New Graph Topological Signatures},
  booktitle = {Proceedings of the Twenty Third International Conference on Artificial Intelligence and Statistics},
  series    = {Proceedings of Machine Learning Research},
  volume    = {108},
  pages     = {2786--2796},
  year      = {2020},
  publisher = {PMLR}
}

@article{JMLR:v20:18-358,
  author  = {Hofer, Christoph D. and Kwitt, Roland and Niethammer, Marc},
  title   = {Learning Representations of Persistence Barcodes},
  journal = {Journal of Machine Learning Research},
  volume  = {20},
  number  = {126},
  pages   = {1--45},
  year    = {2019}
}
\bibliographystyle{spiebib}

\end{document}